\documentclass[10pt,twocolumn,letterpaper]{article}

\usepackage[pagenumbers]{cvpr} % To force page numbers, e.g. for an arXiv version

\usepackage{tikz}
\usepackage{pgfplots}
\usepgfplotslibrary{groupplots}
\usepackage{graphicx}
\pgfplotsset{compat=1.18}
\usepackage{xcolor}
\usepackage[table]{xcolor}
\definecolor{cvprblue}{rgb}{0.21,0.49,0.74}
\usepackage[pagebackref,breaklinks,colorlinks,allcolors=cvprblue]{hyperref}
\usepackage{adjustbox}
\usepackage{booktabs}
\usepackage{multirow}
\usepackage{graphicx}
\usepackage{amsmath, amssymb}
\usepackage{mathtools}
\usepackage{wrapfig} % preamble에 추가
\usepackage[accsupp]{axessibility}

\usepackage{pifont}
\usepackage{float}

\usepackage[resetlabels]{multibib}
\newcites{sup}{Supplementary References}

\def\paperID{} % *** Enter the Paper ID here
\def\confName{CVPR}
\def\confYear{2026}
\usepackage{pifont}

\usepackage{float} % preamble에 추가
\definecolor{fail}{RGB}{255, 210, 210}   % 연한 핑크
\definecolor{mid}{RGB}{255, 240, 200}    % 살짝 노랑
\definecolor{ok}{RGB}{220, 255, 220}     % 연초록
\definecolor{sky}{RGB}{210, 230, 255}
\title{AcTTA: Rethinking Test-Time Adaptation via Dynamic Activation}

\author{
    Hyeongyu Kim$^{1}$\thanks{Equal contribution} \quad Geonhui Han$^{1}$\footnotemark[1] \quad Dosik Hwang$^{1,2}$\thanks{Corresponding author} \\
    $^1$ Yonsei University 
    \quad $^2$ Korea Institute of Science and Technology 
    \\
    {\tt\small \{lion4309, hgh6945, dosik.hwang\}@yonsei.ac.kr} \\
    \vspace{0.2em}
    \small\url{https://hyeongyu-kim.github.io/actta/}
}

\begin{document}
\maketitle
\begin{abstract}
Test-time adaptation (TTA) aims to mitigate performance degradation under distribution shifts by updating model parameters during inference.
Existing approaches have primarily framed adaptation around affine modulation, focusing on recalibrating normalization layers. This perspective, while effective, overlooks another influential component in representation dynamics: \textbf{\textit{the activation function}}.
We revisit this overlooked space and propose \textbf{\textit{AcTTA}}, an activation-aware framework that reinterprets conventional activation functions from a learnable perspective and updates them adaptively at test time.
AcTTA reformulates conventional activation functions (e.g., ReLU, GELU) into parameterized forms that shift their response threshold and modulate gradient sensitivity, enabling the network to adjust activation behavior under domain shifts.
This functional reparameterization enables continuous adjustment of activation behavior without modifying network weights or requiring source data.
Despite its simplicity, AcTTA achieves robust and stable adaptation across diverse corruptions. Across CIFAR10-C, CIFAR100-C, and ImageNet-C, AcTTA consistently surpasses normalization-based TTA methods.
Our findings highlight activation adaptation as a compact and effective route toward domain-shift–robust test-time learning, broadening the prevailing affine-centric view of adaptation.
\end{abstract}    
\section{Introduction}
\label{sec:intro}

Deep neural networks often suffer substantial performance degradation when deployed in environments that differ from their training distribution \cite{koh2021wilds,hendrycks2019benchmarking,hendrycks2021many}. To address this, test-time adaptation (TTA) has emerged as a paradigm for adapting models during inference without access to source data \cite{tent,ttt,liang2025comprehensive,wang2025search}.
Existing approaches span a wide spectrum, from full or partial fine-tuning of model weights to lightweight, layer-specific updates \cite{iwasawa2021test, zhang2022memo, kimbuffer}. Among them, normalization-based strategies have become the most prevalent, typically adapting models by updating affine parameters and recalibrating running statistics of normalization layers \cite{tent,eata,deyo}. These methods have established normalization adjustment as the dominant mechanism for TTA, under the assumption that aligning feature statistics is the most effective way to mitigate domain shift.

% \begin{figure}[t]
% \centering
% \includegraphics[width=0.8\linewidth]{figs/main_fig1.png}
% \caption{An example visualization of the adaptive activation function. This is an example.}
% \label{fig:activation_example}
% \end{figure}

However, this assumption leaves a fundamental and highly influential component almost untouched: \textbf{\textit{the activation function}}. The nonlinearity introduced by activations fundamentally shapes the geometry of feature spaces and governs how representations respond to input variations \cite{goodfellow2013maxout, agarap2018deep,ramachandran2017searching}. Despite this central role, activation functions have been treated as fixed nonlinear mappings during adaptation, ignoring their potential as adaptable mechanisms for domain alignment.

%Earlier works demonstrated that learnable nonlinearities can flexibly reshape decision boundaries and enhance robustness under complex input distributions, yet such adaptability has rarely been revisited in modern test-time adaptation frameworks. 

Interestingly, outside the TTA context, a growing body of work has explored learnable or modulated activation functions, ranging from parametric forms to gated variants, which enable richer and more adaptive representation dynamics \cite{he2015delving, molina2019pad, ma2021activate, agostinelli2014learning}. These studies consistently show that even subtle modifications in activation behavior can lead to improvements in performance and training stability. This line of work suggests that activation functions inherently possess a learnable degree of variability that can modulate how features respond to input shifts. Yet, this flexibility has remained largely unexplored in the TTA setting, where the role of activation adaptability has been largely overlooked.

Activation functions serve as the nonlinear core of neural networks, shaping how features transform and how gradients propagate.
In the context of test-time adaptation, this nonlinearity plays a critical yet underexplored role: it determines how sensitively the model responds to shifted feature distributions.
If the activation itself could adapt while adjusting its response center or modulating its gradient flow, the network could internally re-balance without altering its weights.
Such adaptability would enable more fine-grained and localized correction of feature drift, complementing existing TTA strategies that typically rely on external normalization or weight updates.
This insight motivates rethinking the activation function as an active participant in adaptation rather than a fixed component.

To explore this idea, we introduce a simple activation-aware TTA framework, termed \textbf{AcTTA}, which modulates the parameters shaping the activation shift and gradient flow.
By centering adaptation on these activation dynamics, AcTTA allows the model to refine its nonlinear responses to distribution shifts while keeping the overall architecture largely unchanged.

Our contributions can be summarized as follows:
\begin{itemize}
\item We introduce \textbf{AcTTA}, a new perspective on TTA that leverages adaptive activation modulation as an effective and compact way to handle distribution shifts.
\item We design a \textbf{modular and easily pluggable activation adaptation module} that enables activation behavior to be dynamically adjusted during inference, without source data or architectural modifications.
\item Extensive experiments on CIFAR10-C, CIFAR100-C, and ImageNet-C demonstrate consistent improvements in performance and stability.
\end{itemize}

\section{Related Work}
\label{sec:formatting}

%-------------------------------------------------------------------------
\subsection{Test-Time Adaptation}

Recent advances in TTA have explored diverse axes of adaptation beyond simple normalization updates. One major line of work builds on entropy minimization, encouraging confident and consistent predictions by directly reducing output uncertainty on unlabeled target samples \cite{tent, eata, wang2025search}.
Other approaches emphasize consistency regularization or self-ensembling, maintaining temporal or augmentation-level stability to improve adaptation robustness \cite{cmf, cotta, eata}.
A complementary direction, often referred to as test-time training, leverages auxiliary self-supervised objectives to refine representations during inference \cite{ttt}.
Together, these studies demonstrate that even lightweight, on-the-fly optimization can substantially mitigate domain shift without source data \cite{kimbuffer}.

Despite this growing diversity, most approaches concentrate adaptation on normalization layers or feature statistics, implicitly assuming that aligning affine parameters is the primary means to handle domain shift.
This prevailing focus leaves other influential components such as the activation dynamics that govern nonlinear transformations largely underexplored, motivating our investigation into activation-level adaptation.

%-------------------------------------------------------------------------
\subsection{Dynamic Activation Functions}

Early studies on activation functions primarily sought fixed nonlinearities that improved optimization stability and representational power.
These efforts evolved from early functions such as sigmoid and tanh \cite{lecun2002efficient} to ReLU \cite{nair2010rectified,glorot2011deep} and its modern variants such as Swish \cite{ramachandran2017searching}, ELU \cite{clevert2015fast}, and GELU \cite{hendrycks2016gaussian}, which introduce smoother or self-gated nonlinearities that enhance gradient flow and learning stability.
Within this paradigm, the activation function was regarded as a static architectural choice, serving as an essential yet immutable component of the network.

In contrast, more recent research has explored dynamic activation functions that endow networks with functional flexibility \cite{apicella2021survey}.
Parametric forms such as PReLU \cite{he2015delving} introduced trainable slopes to extend rectifiers, while piecewise-linear and rational formulations like APLU \cite{agostinelli2014learning} allowed networks to learn customized nonlinear mappings from data.
PAU \cite{molina2019pad} represents activations as learnable rational functions, jointly optimizing numerator and denominator coefficients to approximate a wide range of nonlinear behaviors in an end-to-end manner.
ACON \cite{ma2021activate} introduces a gating-based mechanism that dynamically controls whether and how strongly neurons are activated, enabling input-dependent modulation of nonlinearity.
Together, these studies highlight that nonlinear transformations need not remain fixed; they can dynamically adjust and reshape, continuously influencing how representations evolve across domains and conditions.

\vspace{-4pt}
\section{AcTTA}
\label{sec:AcTTA}

\subsection{Dynamic Activation Reparameterization}
\label{sec:method_activation}

Normalization layers such as Batch Normalization (BN) aim to stabilize feature distributions by providing approximately zero-centered and standardized inputs to subsequent activations.
However, under distribution shift, the source statistics used by BN may no longer align with target features, producing biased representations that remain even after normalization.
When such biased features are processed by zero-centered activations like ReLU, Swish, or GELU, informative responses can be suppressed below the activation boundary, leading to information loss and vanishing gradients \cite{mohseni2021shifting}.
We identify this zero-centered rigidity as a key factor limiting adaptation, particularly in TTA settings where feature statistics deviate from the source domain.

Recent works such as ACON \cite{ma2021activate} have relaxed this rigidity by introducing learnable gating functions that preserve nonzero gradients in the negative region.
While such designs mitigate the dead-gradient problem, they still assume a fixed zero-centered boundary and thus cannot effectively handle biased or shifted feature distributions.
In test-time adaptation, where the model must adjust to unseen domains without altering its core parameters, this limitation becomes critical.
We therefore seek a more general activation formulation that (1) adapts its gradient behavior to sustain learning flow, (2) shifts its activation boundary to align with new feature centers, and (3) remains compatible with the source-trained representation, allowing adaptation to emerge naturally from within the activation function itself.

\subsection{AcTTA: Activation Modulation for TTA}
\label{sec:actta_update}

Modern activation functions are locally linear mappings whose slope varies smoothly with the input.
For instance, Swish and GELU can be approximated as $\phi(x) = x \cdot \sigma(\beta x)$, where the sigmoid gate $\sigma(\cdot)$ controls how much of the input passes through.
The derivative,
\[
\phi'(x) = \sigma(\beta x) + \beta x\,\sigma(\beta x)\,(1 - \sigma(\beta x)),
\]
acts as an \textit{input-dependent slope} that transitions continuously between low and high values.
This slope governs gradient propagation, but because it is fixed and zero-centered, it can lead to gradient imbalance or biased updates when feature distributions shift.

To enable adaptive behavior, we explicitly expose this slope as a learnable function:
\[
\lambda(x) = \lambda_{\text{neg}} + (\lambda_{\text{pos}} - \lambda_{\text{neg}})\,\sigma(\beta x),
\]
where $\lambda_{\text{neg}}$ and $\lambda_{\text{pos}}$ represent the asymptotic slopes in the negative and positive regions.
This formulation generalizes the derivative $\phi'(x)$, allowing the gradient scale to adapt asymmetrically across feature dimensions.

However, slope adaptation alone cannot correct feature bias.
Under distribution shift, activations often become misaligned because the zero-centered boundary no longer coincides with the shifted feature mean.
To address this, we introduce a learnable center parameter $c$ that repositions the activation boundary, allowing the activation to re-center itself dynamically according to the target-domain statistics.

Combining these mechanisms yields the final activation:

\begin{equation}
\begin{split}
g(x) = &\;\underbrace{\phi(x - c)}_{\text{shifted base}} \\
&+ \underbrace{\left[\lambda_{\text{neg}}
    + (\lambda_{\text{pos}} - \lambda_{\text{neg}})
    \sigma(\beta(x - c))\right](x - c)}_{\text{adaptive slope modulation}}
\end{split}
\end{equation}

Here, $\phi(\cdot)$ denotes a differentiable base activation (e.g., ReLU or GELU).
When initialized with $\lambda_{\text{neg}}=\lambda_{\text{pos}}=0$ and $c=0$, $g(x)$ exactly recovers the original $\phi(x)$, ensuring compatibility with a pre-trained model. See Supplementary for detailed derivation.

By jointly learning $\lambda_{\text{neg}}$, $\lambda_{\text{pos}}$, and $c$ during adaptation, the activation can locally adjust both its gradient flow and its activation center.
This provides a more expressive and stable adaptation mechanism than conventional normalization layers, which only adjust global affine parameters.
In effect, AcTTA allows the model to self-correct internal biases and preserve gradient dynamics in a domain-adaptive manner, without modifying network weights or accessing source data.

\begin{figure}[t]
\centering
\includegraphics[width=1.0\linewidth]{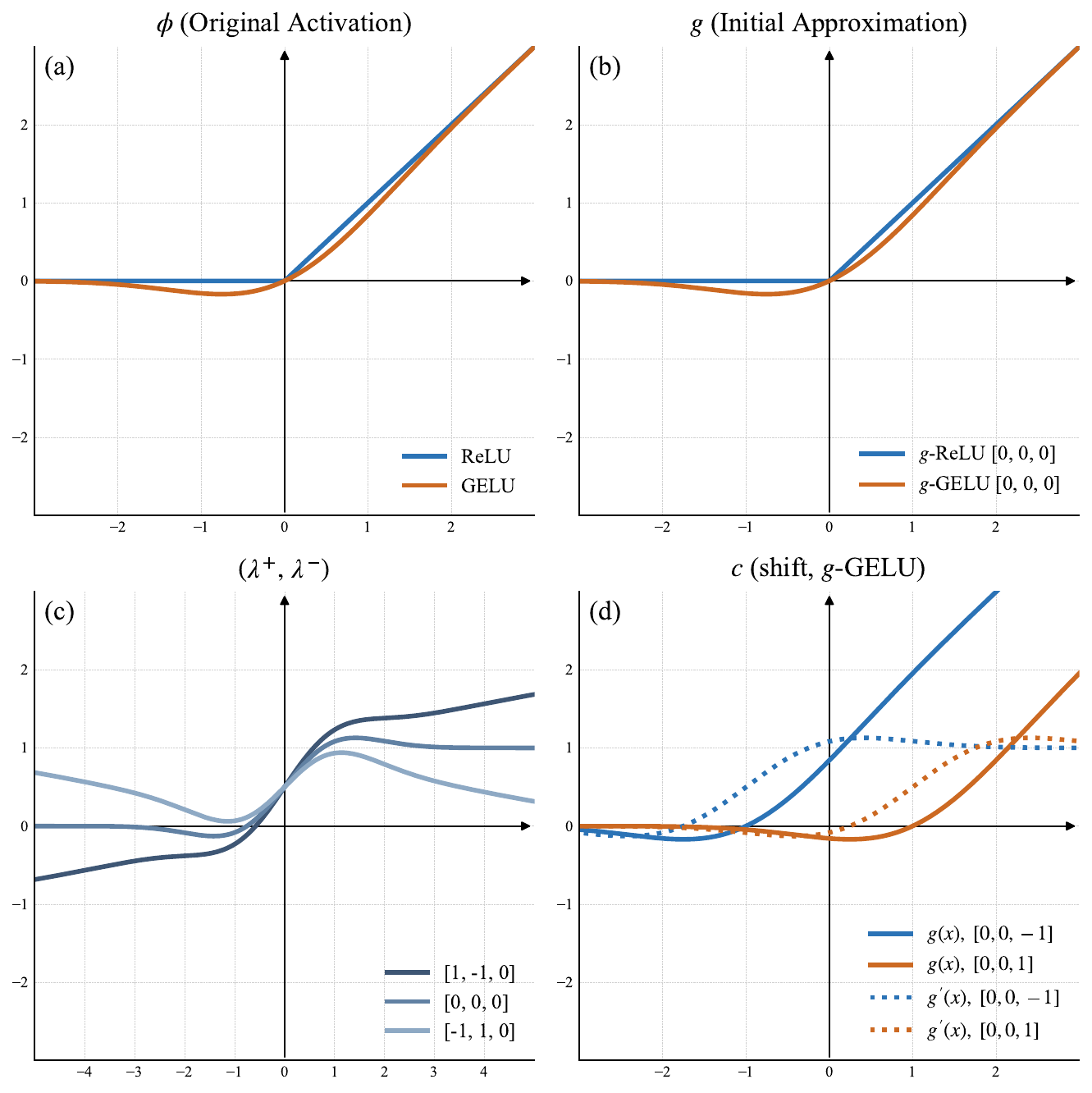}
\caption{
Visualization of the adaptive activation function. 
(a) The base activation functions $\phi(x)$. 
(b) The AcTTA activation output $g(x)$; each parameter set $[\,\cdot,\cdot,\cdot\,]$ denotes 
$[\lambda^{+},\,\lambda^{-},\,c]$.
(c) The first derivative $g'(x)$ with varying $(\lambda^{+},\lambda^{-})$
(d) The effect of the shift parameter $c$, which avoids strict zero-centering during adaptation. 
% \textbf{Color legend.}
% \textcolor{orange}{-----}: ReLU, \;
% \textcolor{cyan}{-----}: GELU, \;
% \textcolor{black}{-----}: AcTTA(ReLU), \;
% \textcolor{red}{-----}: AcTTA(GELU), \;
% \textcolor{ForestGreen}{-----}: AcTTA(GELU)' (derivative).
}
\label{fig:activation_example}
\end{figure}

\vspace{-5pt}

\subsection{Adaptation Objective}
\label{sec:objective}

Most recent TTA studies emphasize \textit{how} the model should be updated, designing complex objectives such as consistency regularization, confidence-based selection, or entropy-weighted filtering to achieve stable adaptation. 
Rather than being tied to a specific update rule, our formulation is motivated by making the activation itself learnable, allowing adaptation to emerge naturally regardless of the employed objective. By simply assigning the learnable target to the activation parameters, the model can self-adjust to distributional shifts without introducing any additional optimization heuristics or auxiliary losses. 
This minimal yet effective objective highlights the modular nature of our approach: 
it can be seamlessly integrated with diverse adaptation strategies while maintaining simplicity and interpretability.

\section{Experiment}
\label{sec:experiment}

\subsection{Implementation Details}
\label{sec:4_1}

We revisit representative TTA methods such as TENT~\cite{tent}, EATA~\cite{eata}, SAR~\cite{sar}, DeYO~\cite{deyo}, ROID~\cite{roid}, and CMF~\cite{cmf} which mainly adapt normalization parameters during inference. For EATA, to comply with the fully source-free assumption, we exclude the Fisher regularization and adopt only the sample-selection ($\rightarrow$ ETA).

Our framework AcTTA is modular and objective-agnostic. 
By making activation parameters learnable alongside affine parameters, it can integrate with existing adaptation objectives (e.g., entropy, consistency, or regularization) without altering their optimization schemes. 
We denote such variants as $\text{AcTTA}_\text{TENT}$, $\text{AcTTA}_{\text{ETA}}$, etc., where the subscript indicates the baseline objective.

\noindent
\textbf{Datasets.} 
Experiments are conducted on CIFAR10-C, CIFAR100-C, and ImageNet-C, each with 15 corruption types, with the highest severity level, which is 5. 
Backbones include WRN-28, ResNeXt, WRN-40, ResNet50 (BN/GN), and ViT-B/16, pretrained on clean training data and adapted online without source access. 
All methods share identical optimization and hyperparameter settings for fair comparison. All results are averaged over three random seeds. Detailed experimental settings, along with additional results on ImageNet-A, ImageNet-R, and PACS, are provided in the supplementary material.

\subsection{Stable Optimization with Larger Learning Rate}
\label{sec:4_2}
Practical TTA, especially in online scenarios, requires the model to adapt immediately as target samples arrive, making fast and responsive updates essential. Although using a large learning rate is desirable for such rapid adaptation, most existing TTA methods easily become unstable or even fail because their gradients vanish in the negative activation region under distribution shift. In contrast, AcTTA preserves nonzero gradients across both positive and negative regions through its adaptive slope design, enabling stable and fast optimization even with aggressive updates. This robustness allows us to safely employ a much larger learning rate; in practice, we use roughly 10× larger than those used in existing TTA methods.

% preamble: 
% \usepackage{booktabs}
% \usepackage{adjustbox}
% \usepackage{xcolor}

\begin{table}[htbp]
\centering
\small
\setlength{\tabcolsep}{2pt}
\renewcommand{\arraystretch}{0.9}

\begin{adjustbox}{max width=\linewidth}
\begin{tabular}{ccccccccc}
\toprule
        & \multicolumn{4}{c}{TENT} & \multicolumn{4}{c}{AcTTA$_\text{TENT}$} \\
\cmidrule(lr){2-5}\cmidrule(lr){6-9}
\textbf{LR / BS} 
        & 4 & 16 & 64 & 128 
        & 4 & 16 & 64 & 128 \\
\midrule
1e-2    
        & 98.51 & 94.41 & 73.28 & 51.57
        & 55.90 & 39.03 & \cellcolor{sky}\textbf{{34.94}} & \cellcolor{sky}\textbf{{34.56}} \\
5e-3    
        & 97.61 & 81.49 & 43.47 & 38.46
        & \cellcolor{sky}\textbf{{55.27}} & \cellcolor{sky}\textbf{{38.87}} & 35.45 &35.31 \\
1e-3    
        & 87.80 & \cellcolor{sky}\textbf{{40.10}} & \cellcolor{sky}\textbf{{35.62}} & \cellcolor{sky}\textbf{{35.25}}
        & 55.66 & 40.39 & 37.74 & 37.61 \\
1e-4    
        & \cellcolor{sky}\textbf{{57.37}} & 41.54 & 38.93 & 38.61
        & - & - & - & - \\
\bottomrule
\end{tabular}
\end{adjustbox}

\caption{
Sensitivity comparison to learning rate and batch size for TENT and AcTTA$_\text{TENT}$ on CIFAR100-C using WRN-40. Sky blue denotes the best performance.
}
\label{tab:lr_sensitivity}
\end{table}

\vspace{-3mm}

\subsection{Trainable Parameter Ablation}
\label{sec:4_3}

Table~\ref{tab:ablation_params} analyzes which components of the proposed activation module are most beneficial to update at test time.
Starting from the Tent~\cite{tent} baseline that adapts only the normalization parameters $(\gamma,\beta)$ with Entropy Minimization (EM),
we progressively introduce the learnable activation parameters of AcTTA:
the activation center $c$ and the asymmetric gradient scales $(\lambda_{\text{pos}}, \lambda_{\text{neg}})$.
On CNNs with BatchNorm (WRN-28), an interesting pattern emerges.
Contrary to the common TTA practice of updating $(\gamma,\beta)$, the best performance is obtained when we \emph{freeze} the BN affine parameters and adapt only the activation parameters $(\lambda_{\text{pos}}, \lambda_{\text{neg}}, c)$ (AcTTA$^\ast$).
Updating $(\gamma,\beta)$ together with the activation parameters slightly degrades accuracy, suggesting that BN and AcTTA play overlapping roles on convolutional backbones.
Since BN already relies on running statistics that are perturbed under shift, further modifying $(\gamma,\beta)$ can amplify distributional noise and overcorrect feature scales.
In contrast, keeping BN fixed while allowing the activation to adjust its center and asymmetric slopes provides a more localized and stable way to compensate residual bias and preserve gradient flow, which is particularly beneficial under our large learning rate setting.

\begin{table}[t]
\centering
\small
\setlength{\tabcolsep}{4pt}
\renewcommand{\arraystretch}{1.05}
\begin{adjustbox}{max width=\columnwidth,center}
\begin{tabular}{lccc}
\toprule
\textbf{Method} & \textbf{Trainable params} & \textbf{WRN-28} & \textbf{ViT-B/16} \\
\midrule
No Adaptation & – & 43.52 & 62.10 \\
Tent   & $\gamma, \beta$ & 18.51 & 53.85 \\
\midrule
AcTTA & $\gamma, \beta, c$ 
& 18.21 & 53.89 \\
AcTTA & $ \gamma, \beta, \lambda_{\text{pos}}$ 
& 18.39 & 53.85 \\
AcTTA & $\gamma, \beta, \lambda_{\text{neg}}$ 
& 18.40 & 52.40 \\
AcTTA & $\gamma, \beta, \lambda_{\text{pos}}, c$ 
& 18.14 & 53.88 \\
AcTTA & $\gamma, \beta, \lambda_{\text{neg}}, c$ 
& 18.13 & 52.44 \\
AcTTA & $\gamma, \beta,\lambda_{\text{pos}}, \lambda_{\text{neg}}$ 
& 18.29 & 52.38 \\
AcTTA & $\gamma, \beta,\lambda_{\text{pos}}, \lambda_{\text{neg}}, c$ 
& 18.06 & \textbf{52.37} \\
\rowcolor{gray!10} AcTTA$_{*}$ & $\lambda_{\text{pos}}, \lambda_{\text{neg}}, c$ 
& \textbf{17.03} & 55.30 \\
\rowcolor{gray!10} AcTTA$_{*}$ & $\lambda_{\text{pos}},  c$ 
& 17.15 & 60.88 \\
\rowcolor{gray!10} AcTTA$_{*}$ & $\lambda_{\text{neg}},  c$ 
& 17.21 & 55.58 \\
\rowcolor{gray!10} AcTTA$_{*}$ & c 
& 17.50 & 61.56 \\

\bottomrule
\end{tabular}
\end{adjustbox}
\vspace{-1mm}

\caption{
Ablation on trainable parameters of AcTTA at test time 
($\lambda_{\text{neg}}$, $\lambda_{\text{pos}}$, and $c$) within the Tent~\cite{tent} framework.
Reported is the error rate on CIFAR10-C (WRN-28) and ImageNet-C (ViT-B/16). 
AcTTA$_{*}$ denotes the variant using a learning rate scaled by $\times 10$ (as defined in Table~1). 
Batch size is 128.
}

\vspace{-2mm}
\label{tab:ablation_params}
\vspace{-2mm}
\end{table}

For ViT-B/16 with LayerNorm, the behavior is noticeably different.
Here, the best configuration is achieved when the LN affine parameters $(\gamma,\beta)$ are adapted jointly with the activation parameters.
Unlike BN, LN normalizes each token on a per-sample basis and does not depend on running statistics from the source domain.
As a result, adapting $(\gamma,\beta)$ in ViTs does not interfere with noisy batch statistics, but instead provides an additional degree of freedom to rescale and re-center features after per-token normalization.
This rescaling is complementary to AcTTA’s activation-level modulation: $(\gamma,\beta)$ adjusts global importance across channels, while $(\lambda_{\text{pos}}, \lambda_{\text{neg}}, c)$ refine the local nonlinear response.
The strong gains observed when both are updated indicate that, under LN, normalization and activation adaptation cooperate rather than compete.

Overall, this ablation highlights that the optimal design of trainable components is backbone dependent.
For BN-based CNNs, it is advantageous to depart from the conventional “update only $(\gamma,\beta)$” paradigm and rely primarily on activation parameters for adaptation.
For LN-based Transformers, jointly adapting normalization and activation proves more effective.
These findings support the modular nature of AcTTA and motivate our default choice of using base AcTTA on CNNs and the full AcTTA configuration on ViTs in subsequent experiments.

A more fine-grained observation concerns the role of the activation-center parameter $c$.
Allowing $c$ to shift already brings noticeable gains, particularly on CNNs with BN, suggesting that residual bias induced by source-domain running statistics can be compensated by adjusting the activation boundary itself. By contrast, Vision Transformers with LN show smaller gains from adapting $c$, since LN normalizes features on a per-sample basis and is less affected by domain-wise mean shift.

%%%%%%%%%%%%%%%%%%%%%%%%%%%%%%%%%%%%%%%%

\subsection{Comparison on Other Learnable Activations}
\label{sec:4_4}

We further compare our adaptive activation formulation against several representative learnable activation functions, 
including PReLU~\cite{he2015delving}, (parametric) GELU~\cite{hendrycks2016gaussian}, ACON~\cite{ma2021activate}, and PAU~\cite{molina2019pad}. 
While these methods introduce parametric flexibility to improve expressiveness during training, 
their structures remain limited to fixed functional forms, which are typically linear or monotonic variants with a small number of learnable coefficients. 
In contrast, our approach parameterizes the activation center and asymmetric slopes, enabling input-dependent reshaping of the response function. 
This formulation allows flexible shift and slope modulation, which is particularly beneficial under distributional shifts encountered at test time.

\vspace{2mm}
\begin{table}[htbp]
\centering
\small
\setlength{\tabcolsep}{6pt} % keep original horizontal spacing
\renewcommand{\arraystretch}{0.95} % tighten ONLY vertical spacing

\vspace{-2mm}

\begin{adjustbox}{max width=\columnwidth,center}
\begin{tabular}{lcc}
\toprule
\textbf{Activation Function} & \textbf{WRN-28} & \textbf{ViT-B/16} \\
\midrule
Original     & 18.51 & 53.85 \\
PReLU / PGELU~\cite{he2015delving}  & 29.62 & 53.87 \\
ACON~\cite{ma2021activate}   & 18.46 & 52.92 \\
PAU~\cite{molina2019pad}     & 34.50 & 99.96 \\
\rowcolor{gray!10}
AcTTA (Ours)  & \textbf{18.06} & \textbf{52.37} \\
\bottomrule
\end{tabular}
\end{adjustbox}

\caption{
Comparison of activation functions under TTA on CIFAR10-C and ImageNet-C under Tent framework.
Results are average classification error (\%) on WRN-28 and ViT-B/16.
}
\label{tab:activation_comparison}
\vspace{-3mm}
\end{table}

\vspace{-1mm}

To ensure a fair comparison, we adopt the same adaptation setting (EM) and backbone architectures 
as in prior sections, replacing the base activation function with each candidate method. 
As summarized in Table~\ref{tab:activation_comparison}, AcTTA delivers strong overall performance across a wide spectrum of datasets and architectural families such as CNN and ViT. Notably, while methods like ACON and PAU provide moderate adaptability through parametric scaling, 
our formulation yields lower error rates in most settings under corruption shifts. 
This suggests that test-time adaptation benefits not merely from parameterized slope control 
but from the joint modulation of the activation center and asymmetric slopes.

In essence, the gain stems from explicitly parameterizing the activation center and asymmetric slopes, allowing the response function to adapt to new domain statistics without overwriting pretrained behavior. This adaptive yet stable mechanism explains why our formulation generalizes better under unseen corruptions than traditional learnable activations with more limited adaptation flexibility.

\subsection{Depth-wise Analysis of Learnable Activations}
\label{sec:4_5}

A natural question arises: would making all activation functions learnable always lead to better adaptation? 
Prior studies often note that domain shifts tend to manifest more strongly in the early feature extraction stages, 
while deeper layers capture more domain-invariant semantics \cite{kimbuffer}. 
Although parameterizing every activation function increases representational flexibility, 
it also introduces additional computational cost and potentially unstable updates during online adaptation. 
Striking a balance between adaptability and efficiency is therefore crucial.

\begin{table}[H]
\centering
\small
\setlength{\tabcolsep}{5pt}
\renewcommand{\arraystretch}{1.1}

\begin{tabular}{cccccc}
\toprule
\textbf{Model}  & \textbf{$\sim$10\%}  & \textbf{$\sim$25\%} & \textbf{$\sim$50\%} & \textbf{$\sim$75\%} & \textbf{$\sim$100\%} \\
\midrule
WRN-40    & \cellcolor{sky}36.17 & \cellcolor{sky}35.14 & \cellcolor{sky}34.05 & 35.13 & 44.97 \\
ResNet50  & \cellcolor{sky}67.67 & \cellcolor{sky}67.02 & \cellcolor{sky}65.89 & \cellcolor{sky}63.45 & 63.47 \\
ViT-B/16  & \cellcolor{sky}56.29 & \cellcolor{sky}51.40 & 52.47 & 57.81 & 65.84 \\
\bottomrule
\end{tabular}

\caption{
Effect of learnable depth ratio in AcTTA across different architectures 
(WRN on CIFAR100-C; ResNet50/ViT on ImageNet-C). 
Sky-blue cells denote ranges showing performance improvement.
}
\label{tab:depth_ablation}
\end{table}

\vspace{-3mm}

To further investigate this trade-off, we vary how deep into the network activations are made learnable. Our results show that the optimal adaptation depth is architecture-dependent: different architectures exhibit distinct sensitivity profiles, making it difficult to identify the best depth a priori. While deeper adaptation provides more flexibility, it also increases computational cost and the risk of unstable updates, creating a trade-off between effectiveness and reliability. Given these constraints and the practical demands of online TTA, we adopt a mid-depth configuration (50\%) as a principled compromise, offering strong and stable performance across architectures without excessive overhead.

\subsection{Baseline Comparison: Large Batch Size}
\label{sec:4_6}

\begin{table*}[htbp]
\centering
\small
\setlength{\tabcolsep}{6pt}
\renewcommand{\arraystretch}{1.05}

\vspace{1mm}

\begin{adjustbox}{max width=\textwidth,center}
% \begin{tabular}{l c @{} c @{} c @{} c @{} c @{} c @{} c @{} c}
\begin{tabular}{lccccccc}

\toprule
\multirow{2}{*}{\textbf{Method}} &
\multicolumn{2}{c}{\textbf{CIFAR10-C}} &
\multicolumn{2}{c}{\textbf{CIFAR100-C}} &
\multicolumn{3}{c}{\textbf{ImageNet-C}}\\

& WRN28 & ResNeXt & WRN40 & ResNeXt & ResNet50 (BN) & ResNet50 (GN) & ViT-B/16 (LN) \\

% \cmidrule(lr){1}\cmidrule(lr){2-3}\cmidrule(lr){4-5}\cmidrule(lr){6-8}

\specialrule{0.1pt}{0pt}{0pt} % midrule보다 얇은 선
Source  
& 43.52  % cifar10 wrn28
& 17.98  % cifar10 resnext
& 46.75  % cifar100 wrn40
& 46.45  % cifar100 resnext
& 82.03  % imagenet resn50bn
& 69.52   % imagenet resn50gn
& 62.10 \\% imagenet vit \\

\specialrule{0.1pt}{0pt}{0pt} % midrule보다 얇은 선

BN   
& 20.61  % cifar10 wrn28
& 13.31  % cifar10 resnext
& 39.51  % cifar100 wrn40
& 35.61  % cifar100 resnext
& 68.10  % imagenet resn50bn
& 69.52  % imagenet resn50gn
& 62.10 \\ % imagenet vit 

\specialrule{0.1pt}{0pt}{0pt} % midrule보다 얇은 선

%%% Tent
TENT~\cite{tent}

& 18.51 ($\pm$0.01)  % cifar10 wrn28
& 10.28 ($\pm$0.01) % cifar10 resnext
& 35.25 ($\pm$0.01) % cifar100 wrn40
& 31.25 ($\pm$0.05) % cifar100 resnext
& 66.50 ($\pm$0.05) % imagenet resn50bn
& 69.60 ($\pm$0.13) % imagenet resn50gn
& 53.85 ($\pm$0.18)  \\ % imagenet vit 

%%% Tent ACTTA
\rowcolor{gray!10}
$\text{AcTTA}_{{\text{TENT}}}$            

& 17.03 ($\pm$0.05)  % cifar10 wrn28
& 9.53 ($\pm$0.03) % cifar10 resnext
& 33.81 ($\pm$0.09) % cifar100 wrn40
& 30.54 ($\pm$0.01) % cifar100 resnext0
& 64.95 ($\pm$0.02) % imagenet res50bn
& 66.84 ($\pm$0.02) % imagenet res50gn
& 51.79 ($\pm$0.02) \\ % imagenet vit

%%% Eata
\specialrule{0.1pt}{0pt}{0pt} % midrule보다 얇은 선
ETA~\cite{eata}

& 18.07 ($\pm$0.38)  % cifar10 wrn28
& 9.74 ($\pm$0.05) % cifar10 resnext
& 36.33 ($\pm$0.01) % cifar100 wrn40
& 31.07 ($\pm$0.02) % cifar100 resnext
& 62.84 ($\pm$0.02) % imagenet resn50bn
& 68.34 ($\pm$0.03) % imagenet resn50gn
& 49.84 ($\pm$0.49)  \\ % imagenet vit 

%%% Eata ACTTA
\rowcolor{gray!10} $\text{AcTTA}_{{\text{ETA}}}$        

& 16.74 ($\pm$0.23)  % cifar10 wrn28
& 9.46 ($\pm$0.03) % cifar10 resnext
& 34.90 ($\pm$0.28) % cifar100 wrn40
& 31.00 ($\pm$0.10) % cifar100 resnext0
& 61.74 ($\pm$0.02) % imagenet res50bn
& 61.37 ($\pm$0.05) % imagenet res50gn
& 48.90 ($\pm$0.30) \\ % imagenet vit

%%% sar
\specialrule{0.1pt}{0pt}{0pt} % midrule보다 얇은 선
SAR~\cite{sar}

& 20.55 ($\pm$0.06)  % cifar10 wrn28
& 12.97 ($\pm$0.00) % cifar10 resnext
& 37.13 ($\pm$0.04) % cifar100 wrn40
& 33.03 ($\pm$0.02) % cifar100 resnext
& 66.24 ($\pm$0.04) % imagenet resn50bn
& 69.37 ($\pm$0.07) % imagenet resn50gn
& 53.90 ($\pm$0.09)  \\ % imagenet vit 

%%% sar ACTTA
\rowcolor{gray!10} $\text{AcTTA}_{{\text{SAR}}}$               

& 20.54 ($\pm$0.01)  % cifar10 wrn28
& 12.95 ($\pm$0.01) % cifar10 resnext
& 35.90 ($\pm$0.11) % cifar100 wrn40
& 32.38 ($\pm$0.02) % cifar100 resnext0
& 64.43 ($\pm$0.02) % imagenet res50bn
& 66.62 ($\pm$0.02) % imagenet res50gn
& 54.17 ($\pm$0.19) \\ % imagenet vit

%%% roid
\specialrule{0.1pt}{0pt}{0pt} % midrule보다 얇은 선
ROID~\cite{roid}

& 17.52 ($\pm$0.06)  % cifar10 wrn28
& 9.62 ($\pm$0.05) % cifar10 resnext
& 35.38 ($\pm$0.06) % cifar100 wrn40
& 30.12 ($\pm$0.13) % cifar100 resnext
& 59.17 ($\pm$0.09) % imagenet resn50bn
& 64.39 ($\pm$0.05) % imagenet resn50gn
& 52.91 ($\pm$0.11)  \\ % imagenet vit 

%%% roid ACTTA
\rowcolor{gray!10} $\text{AcTTA}_{{\text{ROID}}}$             
 
& 16.43 ($\pm$0.08)  % cifar10 wrn28
& 9.43 ($\pm$0.01) % cifar10 resnext
& 33.59 ($\pm$0.01) % cifar100 wrn40
& \textbf{29.49} ($\pm$0.26) % cifar100 resnext0
& 59.74 ($\pm$0.08) % imagenet res50bn %5e-3 range3
& 55.87 ($\pm$0.02) % imagenet res50gn
& 49.89 ($\pm$0.26) \\ % imagenet vit

%%% cmf
\specialrule{0.1pt}{0pt}{0pt} % midrule보다 얇은 선
CMF~\cite{cmf}

& 17.36 ($\pm$0.08)  % cifar10 wrn28
& 9.67 ($\pm$0.05) % cifar10 resnext
& 35.27 ($\pm$0.07) % cifar100 wrn40
& 30.05 ($\pm$0.10) % cifar100 resnext
& \textbf{59.06} ($\pm$0.05) % imagenet resn50bn
& 64.30 ($\pm$0.09) % imagenet resn50gn
& 50.85 ($\pm$0.49)  \\ % imagenet vit 

%%% cmf ACTTA
\rowcolor{gray!10} $\text{AcTTA}_{{\text{CMF}}}$

& \textbf{16.39} ($\pm$0.44)  % cifar10 wrn28
& \textbf{9.41} ($\pm$0.01) % cifar10 resnext
& 33.57 ($\pm$0.12) % cifar100 wrn40
& 29.62 ($\pm$0.20) % cifar100 resnext0
& 59.95 ($\pm$0.02) % imagenet res50bn
& \textbf{55.46} ($\pm$0.31) % imagenet res50gn
& 50.44 ($\pm$0.02) \\ % imagenet vit

%%% deyo
\specialrule{0.1pt}{0pt}{0pt} % midrule보다 얇은 선
DeYO~\cite{deyo}              
& 18.31 ($\pm$0.07)  % cifar10 wrn28
& 9.73 ($\pm$0.08) % cifar10 resnext
& 35.17 ($\pm$0.04) % cifar100 wrn40
& 30.28 ($\pm$0.14) % cifar100 resnext
& 62.34 ($\pm$0.05) % imagenet resn50bn
& 69.02 ($\pm$0.13) % imagenet resn50gn
& \textbf{47.73} ($\pm$0.30)  \\ % imagenet vit 

%%% deyo ACTTA
\rowcolor{gray!10} $\text{AcTTA}_{{\text{DeYO}}}$

& 17.11 ($\pm$0.09)  % cifar10 wrn28
& 9.56 ($\pm$0.01) % cifar10 resnext
& \textbf{33.55} ($\pm$0.53) % cifar100 wrn40
& 30.04 ($\pm$0.72) % cifar100 resnext0
& 61.79 ($\pm$0.04) % imagenet res50bn
& 63.36 ($\pm$0.24) % imagenet res50gn
& 49.64 ($\pm$0.19) \\ % imagenet vit

\bottomrule
\end{tabular}
\end{adjustbox}

\caption{Comparison with state-of-the-art TTA methods on \textbf{CIFAR10-C}, \textbf{CIFAR100-C}, and \textbf{ImageNet-C} (severity level 5). Bold numbers indicate the best results in each column.
}
\label{tab:tta_cifar_batch}
\vspace{-3mm}
\end{table*}

\noindent

As summarized in Table~\ref{tab:tta_cifar_batch}, AcTTA provides consistent and substantial improvements over state-of-the-art TTA methods across a wide spectrum of datasets and architectural families.  
A key strength of our framework lies in its \textbf{high modularity}: activation adaptation is introduced as a plug-in and objective-agnostic component, allowing seamless integration with diverse baseline strategies.
Despite the heterogeneous nature of these baselines, each relying on distinct optimization dynamics and statistical assumptions, AcTTA enhances their performance without requiring any modification to their core objectives or update rules. This confirms that activation-level modulation acts as a broadly compatible mechanism that complements, rather than competes with, existing adaptation pipelines.

Across CIFAR10-C, CIFAR100-C, and ImageNet-C, AcTTA reduces error rates in most evaluated architectures, including WRN, ResNeXt, ResNet50 with both BN and GN, and ViT-B/16 with LayerNorm. The breadth of these improvements highlights the generality of our formulation.  
By endowing the network with adaptive activation mechanisms, AcTTA enables a more flexible functional response that can absorb distribution shifts independently of the underlying normalization scheme.  
Although the magnitude of improvement varies across baselines, AcTTA generally yields positive gains in most settings. This reinforces that activation adaptation introduces a complementary capability that operates orthogonally to normalization-based updates, providing a robust and architecture-agnostic avenue for enhancing test-time performance.

Notably, the performance lift is most pronounced on ImageNet-C, where distribution shifts are more complex and high-capacity models tend to amplify instability during online adaptation.  
The fact that AcTTA maintains robustness in these challenging settings indicates that its modulation of activation functions yields a more structurally reliable adaptation path than relying solely on normalization parameters.  
Together, these findings validate the effectiveness and versatility of activation-aware test-time adaptation, establishing AcTTA as a practical and scalable enhancement to a wide range of TTA methodologies.

\subsection{Baseline Comparison: Small Batch Size}
\label{sec:4_7}

\begin{table}[htbp]
\centering
\small
\setlength{\tabcolsep}{6pt}
\renewcommand{\arraystretch}{1.05}
\setlength{\aboverulesep}{0pt}
\setlength{\belowrulesep}{0pt}
\setlength{\extrarowheight}{-0.2pt}

\begin{tabular}{lccc}
\toprule
\multirow{2}{*}{\textbf{Method}} &
\multicolumn{1}{c}{\textbf{CIFAR10-C}} & \multicolumn{1}{c}{\textbf{CIFAR100-C}} & \multicolumn{1}{c}{\textbf{ImageNet-C}} \\
\cmidrule(lr){2-4}
& WRN28 & WRN40 & ViT-B/16 \\
\midrule
Source            & 43.52  & 46.75 & 62.10 \\

\specialrule{0.1pt}{0pt}{0pt} %
TENT~\cite{tent} & 30.36  & 57.35  & 89.92   \\
 
\rowcolor{gray!10} $\text{AcTTA}_{{\text{TENT}}}$  
& 29.56  & \textbf{55.03}  & \textbf{62.07}   \\

\specialrule{0.1pt}{0pt}{0pt} % midrule보
ETA~\cite{eata} & 30.09 & 57.49  & 98.37  \\

\rowcolor{gray!10} $\text{AcTTA}_{{\text{ETA}}}$  
& 29.11  & 57.04  & 72.70   \\

\specialrule{0.1pt}{0pt}{0pt} % midrule보
SAR~\cite{sar}  & 31.49  & 57.20  & 71.81   \\

\rowcolor{gray!10} $\text{AcTTA}_{{\text{SAR}}}$  
& 31.03  & 56.93  & {82.11}   \\

\specialrule{0.1pt}{0pt}{0pt} % midrule보
ROID~\cite{roid} & 30.87  & 58.75  & 99.80  \\

\rowcolor{gray!10} $\text{AcTTA}_{{\text{ROID}}}$  
& 29.62 & 56.38  & 86.37   \\
\specialrule{0.1pt}{0pt}{0pt} % midrule보

CMF~\cite{cmf} & 30.19  & 58.27  & 99.79   \\

\rowcolor{gray!10} $\text{AcTTA}_{{\text{CMF}}}$  
& \textbf{28.64}  & 55.72  & 87.08  \\

\specialrule{0.1pt}{0pt}{0pt} % midrule보
DeYO~\cite{deyo} & 30.23  & 56.89  & 99.45   \\

\rowcolor{gray!10} $\text{AcTTA}_{{\text{DeYO}}}$  
& 29.25  & 55.14  & 79.49   \\

\bottomrule
\end{tabular}

\caption{Comparison with state-of-the-art TTA methods on small batch size regime (batch size =4). Reported are average classification errors (\%), averaged over 3 runs. Bests are in bold.}
\label{tab:tta_smallbatch}
\vspace{-1mm}
\end{table}

%====================== tablex.tex ======================

\begin{table*}[htbp]
\centering
\small
\setlength{\tabcolsep}{3pt}
\renewcommand{\arraystretch}{0.95}

\begin{adjustbox}{max width=\textwidth}
\begin{tabular}{ll*{15}{c}c}
\toprule
  &                       & \multicolumn{15}{c}{\textbf{Time} $\longrightarrow$}                                                  &        \\
\cmidrule(lr){3-17}
  & \textbf{Method}       & \rotatebox{70}{gaussian} & \rotatebox{70}{shot} & \rotatebox{70}{impulse} & \rotatebox{70}{defocus} & \rotatebox{70}{glass} & \rotatebox{70}{motion} & \rotatebox{70}{zoom} & \rotatebox{70}{snow} & \rotatebox{70}{frost} & \rotatebox{70}{fog} & \rotatebox{70}{brightness} & \rotatebox{70}{contrast} & \rotatebox{70}{elastic} & \rotatebox{70}{pixelate} & \rotatebox{70}{jpeg}  & Mean \\
\midrule

%====================== TARGET BLOCK ======================
\multirow{5}{*}{\rotatebox{90}{\textbf{Target}}}
  & No Adaptation                & 65.69 & 60.05 & 59.10 & 32.04 & 50.96 & 33.57 & 32.35 & 41.39 & 45.17 & 51.38 & 31.64 & 55.51 & 40.24 & 59.70 & 42.43 & 46.75\\
  & BN Adapt              & 45.28 & 44.17 & 47.58 & 32.32 & 46.21 & 32.97 & 33.00 & 39.05 & 38.53 & 45.46 & 30.39 & 36.70 & 40.84 & 37.19 & 43.99 & 39.58 \\
  
  & TENT-continual        & 40.31 & \textbf{37.29} & 41.49 & 33.55 & 43.52 & 35.50 & 32.89 & 39.20 & 38.45 & 42.91 & 36.26 & 41.63 & 47.15 & 45.67 & 57.28 & 40.87 \\
  & CoTTA                 & 43.75 & 41.86 & 43.72 & 32.81 & 43.04 & 33.15 & 33.06 & 38.10 & 37.59 & 45.54 & 29.90 & 37.76 & 39.27 & 33.86 & 39.91 & 38.22\\
  
  \rowcolor{gray!10} \cellcolor{white}\textbf{}  & \textbf{AcTTA$_\text{TENT}$}  & \textbf{39.42} & 37.40 & \textbf{41.20} & \textbf{30.29} & \textbf{42.96} & \textbf{30.52} & \textbf{30.31} & \textbf{34.95} & \textbf{33.58} & \textbf{37.21} & \textbf{28.39} & \textbf{29.57} & \textbf{37.23} & \textbf{31.29} & \textbf{39.05} & \textbf{34.89}\\
\midrule

%====================== SOURCE BLOCK ======================
\multirow{4}{*}{\rotatebox{90}{\textbf{Source}}}
  & BN Adapt              & 27.69 & \textbf{27.37} & \textbf{27.49} & 27.84 & 27.58 & 27.69 & 27.52 & 27.63 & 27.63 & 27.55 & 27.56 & 27.46 & 28.03 & 27.49 & \textbf{27.47} & 27.60 \\  
  & TENT-continual        & \textbf{26.65} & 27.88 & 30.22 & 29.28 & 30.53 & 30.27 & 29.82 & 30.75 & 31.24 & 33.67 & 33.73 & 35.17 & 38.93 & 41.08 & 48.40 & 33.17\\
  & CoTTA                 & 34.31 & 35.06 & 36.02 & 34.59 & 34.73 & 35.12 & 34.23 & 34.09 & 34.05 & 33.39 & 33.49 & 34.65 & 33.65 & 34.34 & 34.05 & 34.38 \\
  \rowcolor{gray!10}  \cellcolor{white}\textbf{} & \textbf{AcTTA$_\text{TENT}$}  &  28.40 & 29.08 & 28.15 & \textbf{25.56} & \textbf{25.97} & \textbf{25.12} & \textbf{24.78} & \textbf{25.67} & \textbf{26.16} & \textbf{26.78} & \textbf{25.04} & \textbf{25.32} & \textbf{25.11} & \textbf{25.63} & 28.32 & \textbf{26.34} \\
\bottomrule
\end{tabular}
\end{adjustbox}

\caption{Error rate comparison (\%) on \textbf{CIFAR100-C} corruptions over time under the continual test-time adaptation setting. Bold numbers indicate the best results in each column.}
\vspace{-3mm}
\label{tab:time_target_source}
\end{table*}

%====================== end of tablex.tex ======================

As shown in Table~\ref{tab:tta_smallbatch}, AcTTA remains effective even under the small-batch regime (batch size = 4), where test-time adaptation is typically more unstable due to noisy gradient estimates and unreliable normalization statistics. Despite these challenges, our activation-based adaptation consistently outperforms baseline methods across CIFAR10-C, CIFAR100-C, and ImageNet-C. This robustness highlights that activation modulation remains an effective adaptation mechanism even when statistical signals are severely limited, a scenario where normalization-based approaches often degrade sharply.

It is important to note that all baseline methods are evaluated using their optimal learning rates, as identified in Table~\ref{tab:lr_sensitivity} and validated for fairness. Nevertheless, AcTTA provides generally positive gains across diverse architectures and objectives. This result reinforces that the advantage of our approach does not arise from hyperparameter tuning or favorable optimization settings, but from the intrinsic stability and flexibility of activation-level adaptation. The ability to deliver reliable improvements under both large- and small-batch conditions further demonstrates the practical applicability of AcTTA in real-world TTA scenarios, where batch sizes are often constrained by latency or memory limits.

\subsection{Continual Forgetting}

\label{sec:4_8}

While continuous test-time adaptation is often regarded as a key requirement for deployment in dynamic real-world environments, we argue that an equally fundamental yet under-explored challenge lies in \emph{continual forgetting}. Existing TTA works predominantly focus on achieving smooth adaptation as the target distribution evolves, and typically measure success only in terms of target-domain accuracy. However, repeated and accumulated updates on non-stationary target streams can gradually overwrite the knowledge acquired from the source domain, causing degradation in performance. This issue is particularly severe in online and long-horizon settings, where access to source data is prohibited and no explicit mechanism exists to recover from overfitting to transient target distributions.

From this perspective, we advocate evaluating TTA methods not only by how well they adapt to continuous domain shifts, but also by how effectively they \textbf{preserve source-domain competence} during adaptation. Our activation-based framework directly addresses this trade-off: AcTTA performs strong adaptation on continuously evolving corruptions while exhibiting substantially reduced forgetting of the pretrained behavior. By enabling flexible yet stable modulation of activation responses, our method maintains robust performance on both source and target domains over extended adaptation horizons. We therefore posit that \emph{continual forgetting} should be treated as a primary diagnostic for practical TTA, and that activation-aware adaptation offers a principled avenue toward mitigating this phenomenon.

A deeper examination reveals that continual forgetting is not merely a side effect of unstable optimization, but a structural vulnerability of existing TTA pipelines. Methods that update normalization statistics or affine parameters at every step can inadvertently bias the representation toward the current mini-batch, especially under severe corruptions or rapidly changing target streams. Since these updates accumulate without any mechanism for reversibility, the model progressively drifts away from the source-trained manifold, resulting in catastrophic degradation when the corruption pattern fluctuates or returns to a distribution closer to the source. In contrast, AcTTA limits adaptation to the activation space, enabling the model to adjust its nonlinear response characteristics without perturbing the core representational geometry encoded by the pretrained weights. This design inherently constrains how much the model can deviate from its original semantics, mitigating long-term drift while still enabling meaningful adaptation. Consequently, AcTTA achieves a more stable equilibrium between reactivity and retention, making it well-suited for long-horizon, real-world deployment scenarios where distribution shifts evolve continuously and unpredictably.

\subsection{Dynamic Activation Behavior Analysis}

\label{sec:4_9}

\begin{figure}[htbp]
\centering
\includegraphics[width=1.0\linewidth]{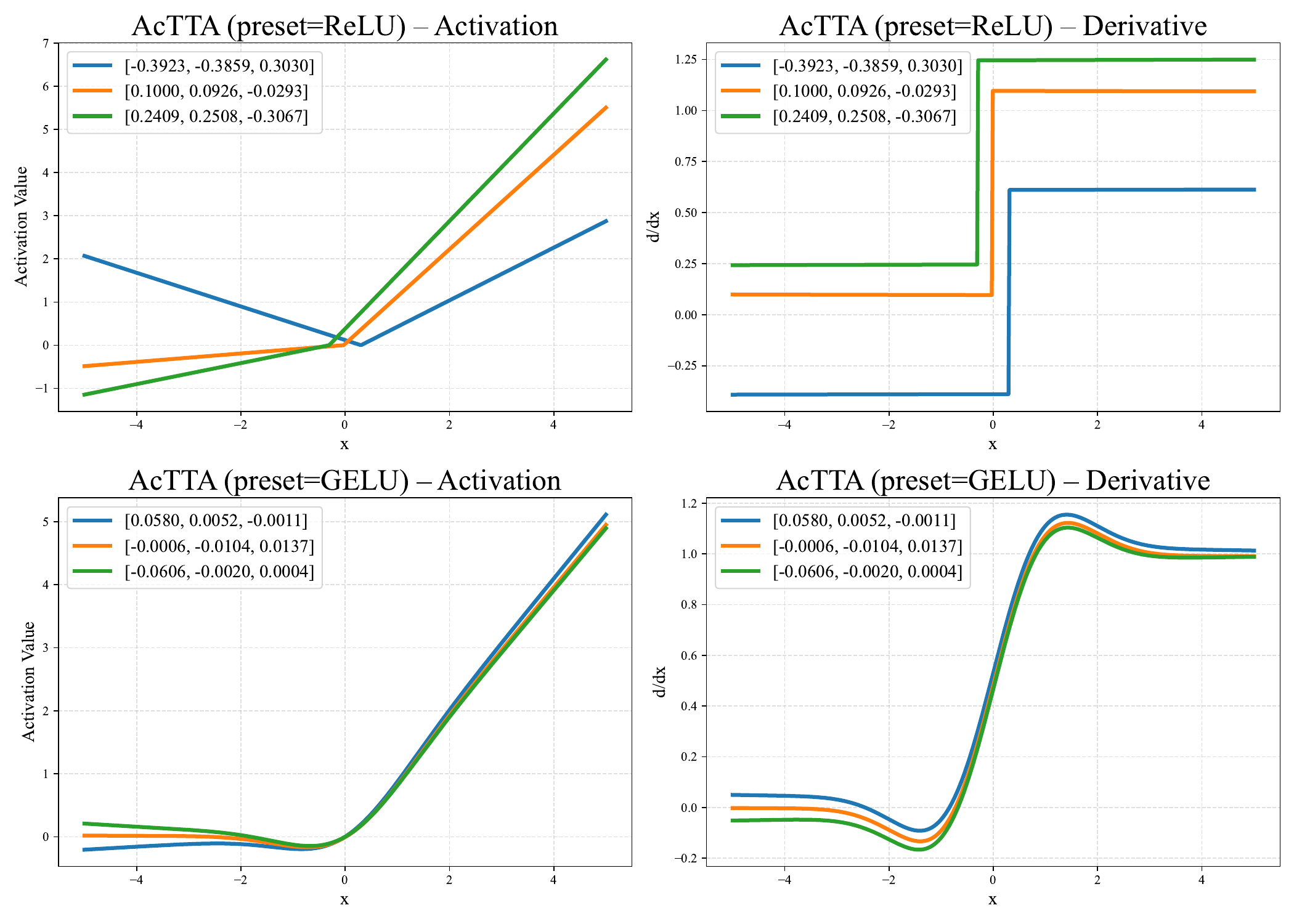}
\caption{Visualizations of adapted activations during AcTTA. 
AcTTA-ReLU results are obtained from WRN-28 on CIFAR10-C after completing adaptation on the Gaussian noise corruption. 
AcTTA-GELU results are from ViT-B/16 on ImageNet-C at the same adaptation endpoint. 
The learned activation exhibits flexible, channel-specific shaping behavior, adapting its form according to the input distribution. The three legend values denote the \textbf{adapted} parameter values of AcTTA, $[\lambda^{+},\,\lambda^{-},\,c]$, corresponding to positive/negative slope modulation and shift.
}

\label{fig:activation_example}
\end{figure}

To better understand the effect of AcTTA during adaptation, we analyze how the activation functions themselves evolve after test-time learning. Figure~\ref{fig:activation_example} visualizes several representative activations extracted from the adapted model. We observe that the learned activation boundaries shift noticeably compared to the source-trained counterparts, confirming that AcTTA effectively recenters the nonlinear response to match the target-domain feature statistics. At the same time, the asymmetric slope parameters adjust to preserve nonzero gradients in regions where conventional activations would typically saturate. These structural changes are not manually imposed but emerge naturally through the optimization of $(\lambda_{\text{pos}}, \lambda_{\text{neg}},c)$ during online adaptation.

Interestingly, the adapted activations differ across layers: early layers tend to exhibit larger center shifts, compensating for distributional biases introduced by corrupted inputs, whereas deeper layers mainly refine gradient scales to stabilize optimization. This layer-specific behavior suggests that AcTTA provides a flexible mechanism that automatically allocates adaptation capacity where it is most needed, without modifying the network weights or relying on source data. Overall, these observations offer concrete evidence that activation-level reparameterization forms a meaningful internal adjustment pathway, enabling the network to realign its representation space and maintain stable gradient flows under challenging target-domain shifts.

\subsection{Resource-Performance Trade-off}

\label{sec:4_10}

Table~\ref{tab:table_memory} presents the resource–performance trade-off of our method compared to existing TTA approaches. Although AcTTA introduces additional learnable activation parameters and maintains gradient flow through them, resulting in a moderate increase in memory usage, it remains substantially more efficient than recent complex TTA algorithms such as CMF and ROID, both of which incur significantly higher VRAM and computational overhead. Despite its lightweight design, AcTTA achieves the lowest error rate among all compared methods, demonstrating that activation-level adaptation offers a highly favorable balance between computational cost and adaptation effectiveness. This confirms that meaningful improvements in TTA robustness do not require heavy architectural modifications or memory-intensive consistency pipelines; instead, fine-grained modulation of activation dynamics provides a compact yet powerful mechanism for reliable adaptation.

\begin{table}[htbp]
\centering
\small
\setlength{\tabcolsep}{6pt}
\renewcommand{\arraystretch}{1.15}

\begin{adjustbox}{max width=\columnwidth,center}
\begin{tabular}{lccc}
\toprule
\textbf{Method} & \textbf{VRAM (MiB)} &  \textbf{Time (ms)} & \textbf{Error (\%)} \\
\midrule
TENT \cite{tent}  & 4013 & 112.83 & 18.51 \\
ROID \cite{roid}  & 9785 &  208.01 & 17.52\\
CMF  \cite{cmf} & 10439 & 218.21  & 17.36 \\

\rowcolor{gray!10}
AcTTA$_\text{TENT}$ (Ours)  & 6249 & 164.51 & \textbf{17.03} \\
\bottomrule
\end{tabular}

\end{adjustbox}
\caption{
Computational efficiency evaluation. 
Experiments were conducted on CIFAR10-C using WRN-28 with a batch size of 128. 
Reported VRAM corresponds to the peak memory usage. 
AcTTA$_{\text{TENT}}$ incurs higher memory and runtime than TENT, but attains the lowest error, outperforming ROID and CMF while remaining substantially more efficient than both.}
\label{tab:table_memory}
\vspace{-5mm}
\end{table}

\section{Conclusion}

We presented AcTTA, a new activation-based perspective on TTA. By parameterizing the activation center and asymmetric slopes, AcTTA mitigates feature bias and preserves gradient flow under distribution shift, while remaining compatible with diverse adaptation objectives.

Experiments on various benchmark datasets show that AcTTA delivers strong overall performance across CNNs and Transformers. Depth-wise analyses show that activation-level adaptation offers favorable trade-offs between robustness and efficiency, while remaining effective under challenging conditions such as small batches, aggressive learning rates, and long-horizon continual adaptation.

AcTTA achieves a strong memory--performance trade-off compared with optimization-heavy adaptation methods, making it a practical alternative to normalization-centric approaches. Beyond its empirical effectiveness, AcTTA offers a new perspective on test-time adaptation by using activation-function modulation as a primary mechanism for online adaptation. At the same time, AcTTA has a clear limitation: its effectiveness depends on design choices such as backbone architecture, activation type, and adaptation depth, and these factors are not yet determined in a fully principled manner. This limits the universality of the method and highlights the need for more systematic selection strategies in future work.

\clearpage \noindent \textbf{Acknowledgements. } This work was supported by the National Research Foundation of Korea (NRF) grant funded by the Korea government (MSIT) (No. RS-2025-16070382, RS-2025-02215070, RS-2025-02217919), Artificial Intelligence Graduate School Program at Yonsei University (RS-2020-II201361), the Korea Institute of Science and Technology (KIST) Institutional Program under Grant 26E0170. 
{
    \small
    \bibliographystyle{ieeenat_fullname}
    \bibliography{main}
}

%%%%% WARNING: do not forget to delete the supplementary pages from your submission 
% \clearpage
\appendix
% \clearpage
% \setcounter{page}{1}
\maketitlesupplementary

\section*{A. Implementation Details}

All TTA experiments were implemented using a public benchmark repository on GitHub, \url{https://github.com/mariodoebler/test-time-adaptation}.

\paragraph{Optimizer and Learning Rate Settings.}
The optimizer and learning rate configurations used for each dataset and model are as follows:
CIFAR-10 and CIFAR-100 experiments use the Adam optimizer with a learning rate of $1\times10^{-3}$.
For ImageNet with ResNet-50, we use SGD with momentum $0.9$ and a learning rate of $2.5\times10^{-4}$ (applied to both BN and GN variants).
For ImageNet with ViT models, we use SGD with momentum $0.9$ and a learning rate of $1\times10^{-3}$. 
For the case of batch size $4$, we scale the base learning rate by a factor of $0.1$ for every method.

\paragraph{Dataset Settings.}
For the corrupted CIFAR datasets (CIFAR10-C and CIFAR100-C), we use all 10{,}000 test images.
For ImageNet, we evaluate on 5{,}000 validation images following standard practice.

\paragraph{Hardware.}
All experiments were conducted using an NVIDIA RTX A6000 GPU and an AMD EPYC 7513 CPU.

%%%%%%%%%%%%%%%%%%%%%%%%%%%%%%%%%%%%%%%%%%%%%%%%%%%%%
%%%%%%%%%%%%%%%%%%%%%%%%%%%%%%%%%%%%%%%%%%%%%%%%%%%%%
%%%%%%%%%%%%%%%%%%%%%%%%%%%%%%%%%%%%%%%%%%%%%%%%%%%%%
%%%%%%%%%%%%%%%%%%%%%%%%%%%%%%%%%%%%%%%%%%%%%%%%%%%%%
%%%%%%%%%%%%%%%%%%%%%%%%%%%%%%%%%%%%%%%%%%%%%%%%%%%%%
%%%%%%%%%%%%%%%%%%%%%%%%%%%%%%%%%%%%%%%%%%%%%%%%%%%%%
%%%%%%%%%%%%%%%%%%%%%%%%%%%%%%%%%%%%%%%%%%%%%%%%%%%%%
%%%%%%%%%%%%%%%%%%%%%%%%%%%%%%%%%%%%%%%%%%%%%%%%%%%%%
%%%%%%%%%%%%%%%%%%%%%%%%%%%%%%%%%%%%%%%%%%%%%%%%%%%%%
%%%%%%%%%%%%%%%%%%%%%%%%%%%%%%%%%%%%%%%%%%%%%%%%%%%%%
% \setcounter{page}{1}
% \maketitlesupplementary

\section*{B. Derivation of the AcTTA Activation}

In this section, we formally derive the proposed AcTTA activation function. We begin by unifying modern activation functions under a common form, analyze their gradient dynamics, and extend them to a learnable, shift-aware formulation suitable for test-time adaptation.

\subsection*{B.1. Unified Representation of Base Activations}

As discussed in prior work \citesup{ma2021activate_sup}, most modern activation functions $\phi(x)$ can be represented as a self-gated function:
\begin{equation}
    \phi(x) \approx x \cdot \sigma(\beta x),
\end{equation}
where $\sigma(\cdot)$ denotes the sigmoid function and $\beta$ is a scaling factor. This formulation unifies several representative non-linearities:
\begin{itemize}
    \item \textbf{ReLU}: Corresponds to $\beta \to \infty$ (in practice, $\beta > 10$).
    \item \textbf{Swish / SiLU}: Corresponds to $\beta = 1$.
    \item \textbf{GELU}: Can be approximated with $\beta \approx 1.702$.
\end{itemize}

\subsection*{B.2. Gradient Perspective of Activation Functions for Adaptation}

Most modern activation functions can be interpreted from two complementary perspectives: a \textbf{gating perspective}, which determines which parts of the representation become active, and a \textbf{gradient perspective}, which captures how information flows backward under that gating pattern. In common piecewise activations such as ReLU, these two perspectives are tightly coupled. Neurons gated “off” suppress their forward signal and simultaneously receive zero gradient, whereas active neurons transmit both activation and gradient. In TTA, however, this coupling becomes particularly restrictive. Since TTA is itself an on-the-fly adaptation process, we posit that the ability to adjust representations in the negative (non-activated) region is crucial. If gradients vanish in this region, the model cannot exploit potentially informative directions for adaptation. Motivated by this insight, we aim to \textbf{control the gradient behavior} independently of the gating structure, enabling \textbf{non-zero and informative gradients even in nominally inactive regions} while retaining the beneficial gating property. This decoupling offers greater flexibility and supports more effective adaptation dynamics during TTA.

\subsection*{B.3. Generalizing the Gradient Behavior}

Let us analyze the derivative of the base function $\phi(x)$:
\begin{equation}
    \phi'(x) = \sigma(\beta x) + \beta x \cdot \sigma(\beta x)(1 - \sigma(\beta x)).
\end{equation}
For practical values of $\beta \ge 1$, the gradient $\phi'(x)$ exhibits specific asymptotic behaviors:
\begin{equation}
    \lim_{x \to -\infty} \phi'(x) = 0, \quad \lim_{x \to +\infty} \phi'(x) = 1.
\end{equation}
This confirms that standard activations act as a fixed gate with restricted bounds: blocking gradients in the negative limit (0) and passing them with a unity scale in the positive limit (1).

To enable learnable adaptation, we require a gradient function whose slope is not restricted to the binary values ${0,1}$ but instead can approach arbitrary optimal limits. Let the desired asymptotic slopes in the negative and positive regions be $\lambda_{neg}$ and $\lambda_{pos}$, respectively. Formally, we seek a generalized slope function $\lambda(x)$ satisfying
\begin{equation}
\lim_{x \to -\infty} \lambda(x) = \lambda_{neg}, \quad
\lim_{x \to +\infty} \lambda(x) = \lambda_{pos}.
\end{equation}

A natural way to achieve this behavior is to start from a smooth gating function that transitions monotonically between two endpoints. The sigmoid $\sigma(\beta x)$ is particularly suitable because it smoothly interpolates from $0$ (as $x \to -\infty$) to $1$ (as $x \to +\infty$). If we interpret this sigmoid output as a continuous interpolation coefficient, then constructing a slope function that moves smoothly between $\lambda_{neg}$ and $\lambda_{pos}$ becomes straightforward.

Following this, the slope at input $x$ is obtained by linearly blending the two endpoints according to the sigmoid gate:
\begin{equation}
\lambda(x)
= \lambda_{neg} + (\lambda_{pos} - \lambda_{neg}) \cdot \sigma(\beta x).
\end{equation}

Thus, the proposed formulation emerges directly from interpreting the sigmoid as a smooth interpolation mechanism. Rather than choosing $\lambda(x)$ arbitrarily, this construction provides the canonical way to transition between two learnable asymptotic slopes while preserving the smooth gating property.

\subsection*{B.4. Derivation via Integral Approximation}

Our objective is to construct a new activation function $g(x)$ that satisfies two conditions:
\begin{enumerate}
    \item \textbf{Backward Compatibility:} It should be able to express the pre-trained baseline $\phi(x)$.
    \item \textbf{Gradient Matching:} The derivative of its residual term should approximate the target slope form $\lambda(x)$.
\end{enumerate}
We formulate $g(x)$ as the base activation plus a learnable residual term:
\begin{equation}
    g(x) = \phi(x) + \int \lambda(x) \, dx.
\end{equation}
However, the exact integral of $\lambda(x)$ involves logarithmic terms (Softplus), which introduces additional computational overhead. Instead, we employ a simple approximation. Consider the term $x \cdot \lambda(x)$. Its derivative is:
\begin{equation}
    \frac{d}{dx}[x \cdot \lambda(x)] = \lambda(x) + x \cdot \lambda'(x).
\end{equation}
In the asymptotic regions ($x \to \pm\infty$), $\lambda(x)$ converges to a constant, causing $\lambda'(x) \to 0$. Consequently, the term $x \cdot \lambda'(x)$ vanishes, and the derivative approximates the target slope:
\begin{equation}
    \frac{d}{dx}[x \cdot \lambda(x)] \approx \lambda(x).
\end{equation}
Thus, we can efficiently approximate the integral term by simply adding $x \cdot \lambda(x)$.

\subsection*{B.5. Shift-Aware Formulation}

Finally, to address the misalignment between the activation function and shifted feature distributions (often caused by normalization statistics), we introduce a learnable center parameter $c$. By substituting $x$ with $(x-c)$, we allow the activation to operate around the target-domain mean.

Combining the shifted base activation with the approximated residual term yields the final AcTTA formulation:
\begin{equation}
    \begin{aligned}
        g(x) &= \phi(x - c) \\
             &\quad + \underbrace{\left[ \lambda_{neg} + (\lambda_{pos} - \lambda_{neg})\sigma\big(\beta(x - c)\big) \right]}_{\lambda(x-c)} \cdot (x - c).
    \end{aligned}
    \label{eq:actta_final}
\end{equation}

Our formulation also provides two important properties. First, it supports an \textbf{identity initialization}: by setting $\lambda_{neg} = \lambda_{pos} = 0$ and $c = 0$, the function reduces to $g(x) = \phi(x)$, ensuring that the model behaves identically to the original network at deployment time. Second, it offers clear \textbf{asymptotic gradient behavior}. As $x \to -\infty$, where $\phi'(x) \to 0$, the effective gradient of $g(x)$ converges to $\lambda_{neg}$; and as $x \to +\infty$, where $\phi'(x) \to 1$, the gradient approaches $1 + \lambda_{pos}$. This guarantees that the network can learn to exploit meaningful gradients in both negative and positive regions, offering dynamic and flexible adaptation behavior.

\section*{C. Comparison of Activation Parameterization Granularity}

\begin{table}[htbp]
\centering
\small
\setlength{\tabcolsep}{6pt}
\renewcommand{\arraystretch}{1.15}

\begin{adjustbox}{max width=\columnwidth,center}
\begin{tabular}{llccc}
\hline
\multirow{2}{*}{Method} & \multirow{2}{*}{Granularity} 
& \multirow{2}{*}{\# param}
& \multicolumn{2}{c}{Error (\%)} \\
\cline{4-5}
& & & BS=4 & BS=128 \\
\hline

\multicolumn{2}{l}{TENT} 
& 17,952 & 30.36 & 18.51 \\
\hline

\multirow{3}{*}{$\text{AcTTA}_{\text{TENT}}$}
& \cellcolor{sky}channel 
& \cellcolor{sky}3,408 
& \cellcolor{sky}\textbf{29.56} 
& \cellcolor{sky}\textbf{17.03} \\
& layer   
& 24   
& 30.16 
& 18.78 \\
& pixel   
& 3,489,792 
& 48.58 
& 58.80 \\
\hline

\end{tabular}
\end{adjustbox}

\caption{Ablation study on the degree of activation flexibility under different batch sizes on CIFAR10-C with a WRN-28 architecture. The best performance is obtained when activation flexibility is controlled on a per-channel basis.}
\label{tab:table_memory_combined}
\end{table}

We investigate how the granularity of activation-function parameterization influences adaptation performance, comparing channel-wise, layer-wise, and pixel-wise configurations. As shown in Table.~\ref{tab:table_memory_combined}, the model exhibits clear differences depending on how the activation flexibility is allocated. Channel-wise activation shows a notable advantage, achieving a lower error rate than both layer-wise and pixel-wise alternatives. This indicates that distributing activation flexibility across channels provides an effective balance between representational adaptability and parameter efficiency. In other words, channel-wise parameterization represents an intermediate point between the structural simplicity of layer-wise modulation and the high representational freedom of pixel-wise modulation, offering a balanced level of parameterization and expressiveness.

% \input{tables/table_supp_memory_2}

% A similar trend is observed when increasing the batch size to 128 in Table~\ref{tab:table_memory_supp2}. Although the overall error naturally decreases thanks to the larger batch statistics, the relative ordering between the parameterization strategies remains consistent. Channel-wise activation yields the best performance, outperforming the baseline TENT formulation as well as the other granularity settings. These results collectively show that channel-level modulation is a robust and scalable choice, offering strong generalization across different batch-size scenarios.

\section*{D. Additional Results on ImageNet-A, ImageNet-R, and PACS}

We further evaluate AcTTA on ImageNet-A, ImageNet-R, and PACS to verify that its effectiveness is not limited to corruption benchmarks. As shown in Table.~\ref{tab:additional_results}, incorporating AcTTA consistently improves three representative TTA baselines across all three datasets. These gains are observed not only on corruption-based benchmarks, but also under more diverse and realistic distribution shifts, including natural adversarial examples in ImageNet-A, rendition shifts in ImageNet-R, and cross-domain shifts in PACS.

\vspace{-3pt}
\begin{table}[htbp]
\centering
\footnotesize
\setlength{\tabcolsep}{4pt}
\renewcommand{\arraystretch}{0.95}
\setlength{\aboverulesep}{0.1pt}
\setlength{\belowrulesep}{0.1pt}
\begin{tabular}{lcccccc}
\toprule
\multirow{2}{*}{} & \multicolumn{2}{c}{TENT} & \multicolumn{2}{c}{ETA} & \multicolumn{2}{c}{DeYO}  \\
% \multirow{2}{*}{} & \multicolumn{2}{c}{ImageNet-R} & DomainNet-126 \\
\cmidrule(lr){2-3}\cmidrule(lr){4-5}\cmidrule(lr){6-7}
Method & Base & AcTTA & Base & AcTTA & Base & AcTTA \\
\cmidrule(lr){1-7}
ImageNet-A & 66.13  & \textbf{59.31} & 63.52 & \textbf{61.33} & 61.27 & \textbf{57.29} \\
ImageNet-R & 48.96  & \textbf{47.18} & 44.55 & \textbf{42.01} & 41.68 & \textbf{41.07} \\
PACS & 16.75  & \textbf{15.14} & 18.20 & \textbf{18.09} & 16.58 & \textbf{13.09}\\
\bottomrule
\end{tabular}
\caption{Additional results on ImageNet-A, ImageNet-R, and PACS.}
\label{tab:additional_results}
\end{table}

These results suggest that the benefit of AcTTA does not depend on a specific corruption type or dataset structure. Rather, activation-function modulation appears to provide a generally useful adaptation mechanism that remains effective across a broad range of test-time distribution shifts. This consistency supports our view that AcTTA is not merely a corruption-specific design, but a more general approach for improving robustness through online activation adjustment during test-time adaptation.

\section*{E. Additional Experiments on Alternative Activations}

We further compare AcTTA with alternative activation functions in Table.~\ref{tab:other_activations} and observe trends consistent with those reported in Table~3 of the main paper. Across both CIFAR10-C and CIFAR100-C, AcTTA consistently achieves the best performance, whereas alternative activation functions often provide limited gains or even cause substantial degradation under test-time adaptation. In particular, PReLU and PAU frequently exhibit severe instability, leading to much higher error rates than the standard baseline in several cases. We emphasize that such behavior is expected and reflects an inherent limitation of directly applying activation functions designed for standard training to the TTA setting.

\vspace{-3pt}
\begin{table}[htbp]
\centering
\small
\setlength{\tabcolsep}{1.5pt}
\renewcommand{\arraystretch}{0.95}
\setlength{\aboverulesep}{0.3pt}
\setlength{\belowrulesep}{0.3pt}
\begin{tabular}{lcccccc}
\toprule
& \multicolumn{3}{c}{CIFAR10-C} & \multicolumn{3}{c}{CIFAR100-C} \\
Method & TENT & ETA & DEYO & TENT & ETA & DEYO\\
\cmidrule{1-1} \cmidrule(lr){2-4} \cmidrule(lr){5-7}
Base  & 18.55  & 18.33 & 18.44 & 31.68 & 31.24 & 30.41 \\
PReLU & 31.73  & 31.56 & 31.61 & 60.97 & 66.53 & 62.70 \\
PAU   & 37.50  & 36.38 & 34.81 & 95.29 & 91.13 & 88.56 \\
ACON  & 18.06  & 17.92 & 18.05 & 30.97 & 31.51 & 30.61 \\
AcTTA & \textbf{17.11} & \textbf{17.00} & \textbf{17.29} & \textbf{30.64} & \textbf{30.99} & \textbf{30.03} \\
\bottomrule
\end{tabular}
\vspace{-2pt}
\caption{Comparison with alternative activation functions. Error rate (\%). Lower is better.}
\label{tab:other_activations}
\vspace{-4pt}
\end{table}

This can be understood from the mismatch between conventional training and TTA. These activation functions were originally developed for \emph{training-time} optimization with abundant data and many update iterations, whereas TTA operates in a considerably more unstable online adaptation regime under distribution shift. As a result, activation designs that are effective during standard training do not necessarily remain reliable in TTA and may instead amplify optimization instability. Moreover, tuning these methods separately for each test distribution would contradict the practical objective of TTA. In addition, some alternatives such as PAU incur substantial computational overhead. Overall, these results suggest that naively transferring training-time activation designs to TTA exposes their inherent limitations, whereas AcTTA is better aligned with the TTA setting and remains robust.

\section*{F. Gradient Flow Analysis}

\begin{wrapfigure}[11]{r}{0.5\linewidth}
    \vspace{-6pt}
    \centering
    \includegraphics[width=\linewidth]{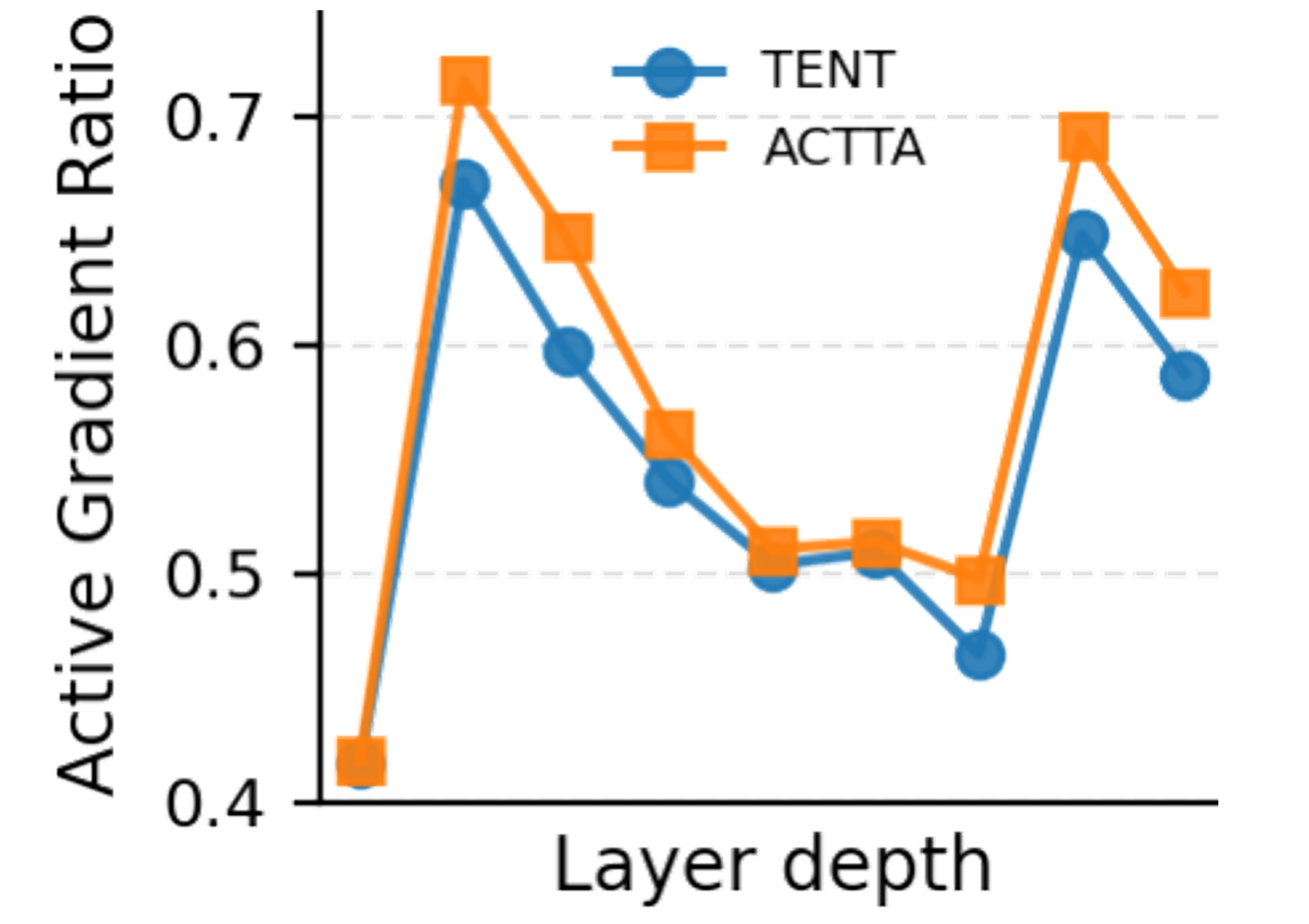}
    \caption{Gradient pass-through ratio.}
    \label{fig_sup_act}
    \vspace{-10pt}
\end{wrapfigure}

We further analyze gradient flow by measuring the layer-wise gradient norms throughout the network.
We also measure the fraction of gradients passing through the activation Figure.~\ref{fig_sup_act}. Compared with TENT using ReLU, AcTTA exhibits consistently larger gradient norms and higher gradient pass-through rates. These observations suggest that AcTTA facilitates gradient propagation during test-time adaptation.

{
    \small
    \bibliographystylesup{ieeenat_fullname}
    \bibliographysup{sup}
}

\end{document}